\documentclass[9pt, conference]{IEEEtran}
\IEEEoverridecommandlockouts
\usepackage{cite}
\usepackage{amsmath,amssymb,amsfonts}
\usepackage{algorithmic}
\usepackage{graphicx}
\usepackage{textcomp}
\usepackage{xcolor}
\usepackage{booktabs}
\usepackage{algorithm}
\usepackage{multirow}
\usepackage{color}
\usepackage{subfigure}
\usepackage{amsmath}
\usepackage{wrapfig}
\usepackage{hyperref}
\usepackage{balance}
\def\BibTeX{{\rm B\kern-.05em{\sc i\kern-.025em b}\kern-.08em
    T\kern-.1667em\lower.7ex\hbox{E}\kern-.125emX}}

\begin{document}

\title{SATformer: Transformer-Based UNSAT Core Learning}

\author{
    \IEEEauthorblockN{Zhengyuan Shi\textsuperscript{1}, Min Li\textsuperscript{1}, Yi Liu\textsuperscript{1}, Sadaf Khan\textsuperscript{1}, Junhua Huang\textsuperscript{2}, Hui-Ling Zhen\textsuperscript{2}, Mingxuan Yuan\textsuperscript{2} and Qiang Xu\textsuperscript{1}}
    \IEEEauthorblockA{
        \textit{\textsuperscript{1} The Chinese University of Hong Kong, \textsuperscript{2} Noah’s Ark Lab, Huawei} \\
        \{zyshi21, mli, yliu22, skhan, qxu\}@cse.cuhk.edu.hk \\
        \{huangjunhua15, zhenhuiling2, Yuan.Mingxuan\}@huawei.com
    }
    \vspace{-20pt}
}

\maketitle

\begin{abstract}
This paper introduces SATformer, a novel Transformer-based approach for the Boolean Satisfiability (SAT) problem. Rather than solving the problem directly, SATformer approaches the problem from the opposite direction by focusing on unsatisfiability. Specifically, it models clause interactions to identify any unsatisfiable sub-problems. Using a graph neural network, we convert clauses into clause embeddings and employ a hierarchical Transformer-based model to understand clause correlation. SATformer is trained through a multi-task learning approach, using the single-bit satisfiability result and the minimal unsatisfiable core (MUC) for UNSAT problems as clause supervision. 
As an end-to-end learning-based satisfiability classifier, the performance of SATformer surpasses that of NeuroSAT significantly. Furthermore, we integrate the clause predictions made by SATformer into modern heuristic-based SAT solvers and validate our approach with a logic equivalence checking task. Experimental results show that our SATformer can decrease the runtime of existing solvers by an average of 21.33\%. 
%In this paper, we propose \textit{SATformer}, a novel Transformer-based solution for Boolean satisfiability (SAT) solving. Different from existing learning-based SAT solvers that learn at the problem instance level, SATformer learns the minimum unsatisfiable cores (MUC) of unsatisfiable problem instances, which provide rich information for the causality of such problems. Specifically, we apply a graph neural network (GNN) to obtain the embeddings of the clauses in the conjunctive normal format (CNF). A hierarchical Transformer architecture is applied on the clause embeddings to capture the relationships among clauses, and the self-attention weight is learned to be high when those clauses forming UNSAT cores are attended together, and set to be low otherwise. By doing so, SATformer effectively learns the correlations among clauses for SAT prediction. Experimental results show that SATformer is more powerful than existing end-to-end learning-based SAT solvers. 
\end{abstract}

\section{Introduction} \label{Sec:Intro}

The Boolean Satisfiability (SAT) problem, fundamental to many fields, seeks to identify if there exists at least one assignment that makes a given Boolean formula \emph{True}. Key applications in the electronic design automation (EDA) field include logic equivalence checking (LEC)~\cite{goldberg2001using}, model checking~\cite{mcmillan2003interpolation}, and automatic test pattern generation (ATPG)~\cite{yang2004trangen}. Being the first proven NP-complete problem, SAT has no complete solution achievable in polynomial time. Most solvers~\cite{sorensson2005minisat,BiereFazekasFleuryHeisinger,selman1993local,cadical,queue2019cadical,fleury2020cadical,audemard2018glucose} use heuristic search techniques for large industrial SAT instances. They return a 'satisfiable' (SAT) result if a valid assignment is found, 'unsatisfiable' (UNSAT) if all search paths yield no valid assignment, or 'unknown' if the maximum runtime is exceeded. Despite impressive results from existing solvers on general benchmarks~\cite{lu2003circuit, audemard2009glucose, audemard2018glucose}, limitations remain in their ability to address EDA problems effectively.

One notable issue is the inability of current solvers to self-direct heuristic selection based on satisfiability. Although modern solvers~\cite{cadical, queue2019cadical, fleury2020cadical, audemard2018glucose} offer satisfiability-specific modes (SAT or UNSAT), they rely on manual selection rather than self-prediction of satisfiability. Moreover, heuristic efficiency can vary markedly when solving SAT and UNSAT instances~\cite{huang2022neural}. For instance, while the local search heuristic can expedite the resolution of SAT instances, it can slow down UNSAT instances, and vice versa for the dynamic restart heuristic. Inappropriate configuration of solver heuristics can potentially increase solving time by up to 50\%~\cite{gagliolo2010algorithm}, as solvers prove SAT and UNSAT in divergent manners. Thus, the prediction of SAT or UNSAT before solving is critical for heuristic adjustment and optimizing the efficiency of heuristic algorithms in solvers.

Another notable limitation of existing solvers is the absence of dedicated heuristics for UNSAT problems. Many EDA applications confront numerous UNSAT problems, which present a greater challenge than SAT problems. For instance, in logic equivalence checking and model checking, a majority of instances are unsatisfiable, necessitating significant runtime to confirm unsatisfiability~\cite{goldberg2003verification,mishchenko2006improvements}. Similarly, in ATPG, proving that some faults are untestable, which corresponds to proving unsatisfiability, consumes a considerable amount of runtime~\cite{liang1995identifying,heragu1997fast}. Hence, integrating an efficient UNSAT heuristic can substantially enhance the generalizability and efficiency of modern solvers within the EDA domain.

%On the other hand, existing solvers lack of dedicated heuristics for UNSAT problems. In practical EDA applications, tools often need to solve numerous UNSAT problems, which are more challenging than SAT problems. For example, in logic equivalence checking and model checking, the majority of instances are unsatisfiable and require significant runtime to prove the unsatisfiability~\cite{goldberg2003verification, mishchenko2006improvements}. Similarly, in ATPG, a substantial amount of runtime is required to prove that some faults are untestable, which corresponds to proving unsatisfiability using a solver~\cite{liang1995identifying, heragu1997fast}. Therefore, incorporating an efficient UNSAT heuristic can significantly improve the generalizability and efficiency of modern solvers in the EDA field. 

To address the above issues, we propose \textit{SATformer}, a novel framework for Boolean satisfiability that not only predicts satisfiability but also accelerates the solving process for unsatisfiable problems. We reinterpret the nature of unsatisfiability as the presence of an unsatisfiable sub-problem and address this by modeling clause interactions. Specifically, the solver asserts unsatisfiability only when an inevitable conflict arises while seeking valid assignments. This conflict is embodied in a subset of clauses known as the Unsatisfiable Core (UNSAT Core), whereby the sub-problem formed by the UNSAT Core is also unsatisfiable. Consequently, by capturing clause interactions and identifying the UNSAT Core, we can determine the satisfiability of the given instance.

To model clause correlation, we formulate a deep learning model that represents clauses as embeddings using a graph neural network (GNN) and discerns clause connections through a novel hierarchical Transformer-based model. This hierarchical model consolidates clauses into clause groups incrementally, capturing relationships between these groups. Hence, our model learns interactions among multiple clauses, transcending the pairwise dependency typical of standard Transformers. Additionally, our model is dual-task trained: for satisfiability prediction with single-bit supervision, and for UNSAT Core prediction supervised by the minimal unsatisfiable core (MUC). Despite an UNSAT instance potentially containing multiple cores of varying size, the core with the minimal number of clauses exhibits less diversity, as highlighted in NeuroCore~\cite{selsam2019guiding}. 

% We further propose a hierarchical Transformer-based model and a multi-task training strategy to model problem instances effectively. On the one hand, since the self-attention mechanism of the standard Transformer only learns pairwise dependency, we design a hierarchical Transformer structure that merges clauses into clause groups level by level and captures the relationships between pairs of clause groups at different scales. On the other hand, our model is trained with two tasks: satisfiability prediction with a single-bit supervision, and the UNSAT Core prediction supervised by the minimal unsatisfiable core (MUC). As noted in NeuroCore~\cite{selsam2019guiding}, the core with the minimal number of clauses has less diversity, despite the possibility of an UNSAT instance containing multiple cores of varying size. Moreover, MUC is the much easier UNSAT sub-problem to prove~\cite{nadel2010boosting}. 

We further deploy SATformer as an initialization heuristic for SAT solvers. SATformer outputs a binary prediction indicating satisfiability and a 'prediction score' for each clause that estimates its likelihood of forming an UNSAT core. We view clauses with high prediction scores as key contributors to unsatisfiability, and propose that addressing sub-problems composed of such high-score clauses can hasten the discovery of UNSAT results. We leverage this clause prediction score to determine search priorities. Specifically, we calculate variable priority scores based on variable-clause connections and use these as the initial variable activity scores, e.g., the Variable State Independent Decaying Sum (VSIDS)~\cite{moskewicz2001chaff}. As most SAT solvers provide a variable branching priority interface~\cite{moskewicz2001chaff, sorensson2005minisat, queue2019cadical, fleury2020cadical}, SATformer can be seamlessly integrated as a plug-in, negating extensive solver reconstruction. Unlike previous variable branching heuristics that solely concentrate on variable activity, our approach utilizes clause-level correlation. We validate the efficacy of SATformer by augmenting the performance of two modern SAT solvers: CaDiCaL~\cite{queue2019cadical} and Kissat~\cite{fleury2020cadical}. 

This work makes the following contributions: 
\begin{itemize}
    \item We propose a fresh perspective on the unsatisfiability problem by modeling clause correlations and identifying the presence of unsatisfiable sub-problems. 
    \item We introduce a hierarchical Transformer-based model to capture clause interactions, leveraging the minimal UNSAT core as supervision, and training our model to distinguish clauses that are \emph{likely} or \emph{unlikely} to form an UNSAT core. 
    \item We integrate our model as a learning-enhanced initialization heuristic into contemporary SAT solvers, yielding a notable acceleration in the solving process. 
\end{itemize}

We organize the rest of the paper as follows. Section~\ref{Sec:Related} reviews the related works about learning-aided SAT solvers and Transformer blocks. We detail our SATformer framework in Section~\ref{Sec:Method} and demonstrate the effectiveness of our model with comprehensive experiments in Section~\ref{Sec:Exp}. Finally, Section~\ref{Sec:Conclusion} concludes this paper.

\section{Related Work} \label{Sec:Related}
\subsection{Deep Learning for SAT Solving}
In the area of combinatorial optimization, the increasing size of problem instances remains a critical challenge. For the SAT problem, deep learning provides an attractive solution to improve the solving efficiency, as surveyed in \cite{guo2022machine}. 

Generally, learning-based SAT solvers fall into two categories: standalone learning-based SAT solvers and learning-aided SAT solvers. On the one hand, standalone learning-based SAT solvers predict satisfiability and decode assignments with end-to-end deep learning models solely. For example, NeuroSAT~\cite{selsam2018learning} treats the SAT problem as a classification task and trains an end-to-end GNN-based framework to predict binary results (satisfiable or unsatisfiable). DG-DAGRNN~\cite{amizadeh2018learning} focuses on the circuit satisfiability problem, i.e., determining whether the single primary output of a given circuit can return logic '1'. They propose a differentiable method to predict a value that evaluates the satisfiaility. By maximizing the value based on reinforcement learning, the model is more likely to find a satisfiable assignment. DeepSAT~\cite{li2022deepsat} formulates the SAT solution as a product of conditional Bernoulli distributions and obtains an assignment by maximizing the joint probability. However, the performance of these standalone learning-based solvers lags behind modern non-learning approaches by a large margin~\cite{selsam2019guiding, li2022deepsat}. Since all assertions produced by deep learning models are based on probabilistic statistic, while SAT problem is a strictly deterministic logic reasoning problem, it is naturally unsuitable to solve the SAT problem with a pure learning method. 

On the other hand, combining deep learning model with modern SAT solvers has emerged as a promising research direction. As heuristics dominate mainstream modern SAT solvers, learning-aided solutions that replace manually-crafted heuristics with more effective and efficient heuristics produced by deep learning models can result in significant performance improvements~\cite{zhang2020nlocalsat, kurin2020can, selsam2019guiding, wang2021neurocomb}. For example, NLocalSAT~\cite{zhang2020nlocalsat} achieves a runtime reduction of $27\% \sim 62\%$ on the stochastic local search (SLS) SAT solver by producing initialization assignments using a deep learning model. Graph-$Q$-SAT~\cite{kurin2020can} replaces the searching heuristic with a reinforcement learning agent and reduces the number of iterations required to solve SAT problems. Moreover, NeuroCore~\cite{selsam2019guiding} and NeuroComb~\cite{wang2021neurocomb} learn the distribution of UNSAT cores to guide SAT solving. Although these learning-aided solutions are powerful, they are generally designed and do not guarantee to speed up a specific kind of problem~\cite{lu2003circuit, audemard2009glucose, audemard2018glucose}. In industrial applications, engineers are often more focused on solving problems in a specific domain. Therefore, a task-specific solution that incorporates prior knowledge about the problems at hand is more practical than a general solution. In this paper, we focus on the unsatisfiable problem in the EDA area and propose a corresponding learning-aided solution. 

\subsection{Transformer and Self-attention Mechanism}
The Transformer~\cite{vaswani2017attention} is well acknowledged as one of the most powerful architectures for modeling sequential data. The adoption of the Transformer has led to tremendous progress in the field of text comprehension~\cite{meng2020readnet}, machine translation~\cite{wang2019learning} and even computer vision~\cite{dosovitskiy2020image,liu2021swin}. The self-attention mechanism in the Transformer block treats sequential input tokens as a fully connected graph and represents the connectivity between every pair of nodes. This enables Transformer-based models to capture the correlation between each pair of tokens. 

The first attempt to involve the Transformer for solving MaxSAT problem (a variant of SAT problem) is introduced in~\cite{shi2021transformer}. The proposed model represents a given SAT instance as a bipartite graph and applies Transformers to aggregate the message over the nodes. In our work, we treat the clauses as a non-ordered sequence and apply the Transformer to capture the correlation among these clauses.

%%%%%%%%%%%%%%%%%%%%%%%%%%%%%%%%%%%%%%%%%%%%%%%%%%%%%%%%%%%%%%%%%%%%%%%%%%%%%%%%%%%%%%
%%%%%%%%%%%%%%%%%%%%%%%%%%%%%%%%%%%%%%%%%%%%%%%%%%%%%%%%%%%%%%%%%%%%%%%%%%%%%%%%%%%%%%

\section{SATformer} \label{Sec:Method}
\subsection{Boolean Satisfiability Problem} \label{Sec:Method:Pre}
A Boolean formula $\phi$ consists of a set of variables $\{x_j\}^I_{j=1}\!\in\!\{True, False\}$ and a set of Boolean operators \{AND ($\land$), OR($\lor$), NOT($\neg$), ...\} over these variables. Solving the SAT problem involves determining whether there is at least one valid assignment of the variables so that the given formula $\phi$ evaluates to $true$. 

Any Boolean formula can be converted into Conjunctive Normal Form (CNF) with an equivalent transformation in linear time~\cite{tseitin1983complexity}. Under CNF conventions, variables and theirs negations are referred to as \textit{literals} (denoted as $x_j$ or $\neg x_j$). A disjunction of several literals constructs a \textit{clause} $C_i = (x_1 \lor x_2 \lor ...)$. A conjunction of clauses forms a propositional instance $\phi = (C_1 \land C_2 \land ...)$. In Eq.~\eqref{eq:sat_example}, we provide an example to exstandard CNF, wherein the instance has three variables $I=\{x_1, x_2, x_3\}$ and consists of three clauses $C_1 = \neg x_1 \lor x_2, C_2 = \neg x_2 \lor \neg x_3, C_3 = x_1 \lor x_3$. 

\begin{equation} \label{eq:sat_example}
\phi := (\neg x_1 \lor x_2) \land (\neg x_2 \lor \neg x_3) \land (x_1 \lor x_3)
\end{equation}

\begin{figure}[!t]
	\centering
	\includegraphics[width=0.55\linewidth]{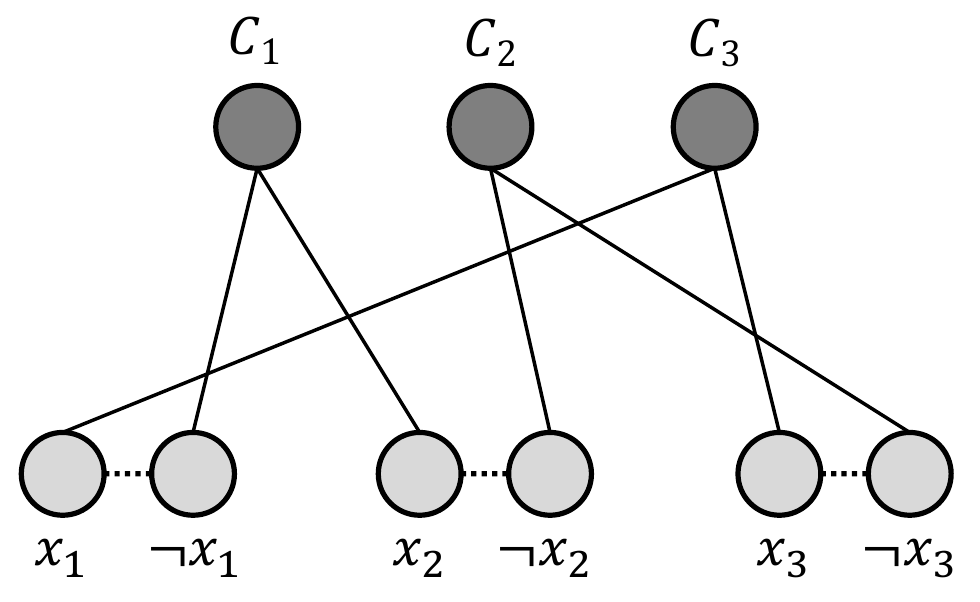}
	\caption{An example of Literal-Clause Graph (LCG).}
	\label{FIG:LCG}
\end{figure}

The Literal-Clause Graph (LCG) is a bipartite graph representation of CNF formulas, which consists of literal nodes and clause nodes. Two types of edges are defined in LCG: one connecting a literal node to a clause node, and another connecting a literal node to its corresponding negation. Fig.~\ref{FIG:LCG} is an example representing Eq.~\eqref{eq:sat_example} in LCG. 

% Since the CNF represents the arbitrary Boolean formula as a normalized combination of two types of nodes, most SAT solving algorithms take CNF as input. Specifically, the conflict-driven clause learning (CDCL) ~\cite{marques2021conflict} based modern SAT solvers~\cite{sorensson2005minisat, BiereFazekasFleuryHeisinger} parse new clauses during backtracking to avoid making the same conflict again. The primary objective of these solvers is searching for an assignment to satisfy all clauses together. The learning-based SAT solver NeuroSAT~\cite{selsam2018learning} and its following works~\cite{selsam2019guiding, ozolins2021goal} apply a GNN on LCG to learn the clause and literal embeddings. In this work, we formulate the SAT instance as a sequence of clauses and then solve the SAT problem by leveraging the correlation among clauses. Therefore, our proposed SATformer also relies on the CNF and its graph representation LCG.

Besides, given an unsatisfiable CNF instance, if a sub-problem constructed by some of its clauses is also unsatisfiable, then such a sub-problem with specific clauses is referred to as an UNSAT core. Take the unsatisfiable instance $\phi_U$ shown in Eq.~\eqref{eq:unsatcore} as an example. We can extract an unsatisfiable sub-problem $\phi_U^{'} = C_2 \land C_5 \land C_6$ as one of the UNSAT cores of the original instance $\phi_U$. The concept of UNSAT core can be further exstandarded by the searching-based solving process. As the clauses $C_2$ and $C_6$ each only include one literal, to satisfy both clauses, the variables must be assigned as $x_1=false$ and $x_2=true$. Nevertheless, this assignment does not satisfy clause $C_5$ under any circumstances, regardless of the value assigned to the remaining variable $x_3$. 
Therefore, the solver proves there is an inevitable \textit{conflict} among these clauses so that all searching branches for solving problem $\phi_U^{'}$ fail. It is worth noting that there is no UNSAT core in a satisfiable instance. 
\begin{equation} \label{eq:unsatcore}
  \begin{split}
    C_1 & = \neg x_1 \lor x_2 \lor \neg x_3 \\
    \underline{C_2} & = x_2 \\
    C_3 & = \neg x_1 \lor \neg x_2 \lor \neg x_3 \\
    C_4 & = \neg x_1 \lor x_2 \lor x_3 \\
    \underline{C_5} & = x_1 \lor \neg x_2 \\
    \underline{C_6} & = \neg x_1 \\
    C_7 & = x_1 \lor \neg x_2 \lor x_3 \\
    C_8 & = x_1 \lor x_2 \lor x_3 \\
    C_9 & = x_1 \lor \neg x_3 \\
    \phi_U & := C_1 \land C_2 ... \land C_9 
  \end{split}
\end{equation}

Since UNSAT cores provide sufficient causality for unsatisfiability judgement, one can determine whether a CNF instance is unsatisfiable or not by extracting its UNSAT cores. In this work, we propose a novel approach to solve the SAT problem by first learning to identify the clauses that contribute to the unsatisfiability of the instance (i.e., clauses included in the UNSAT core) and then modeling the interactions among these clauses.

\subsection{Overview} \label{Sec:Method:Overview}
\begin{figure*}[!t]
	\centering
	\includegraphics[width=0.8\linewidth]{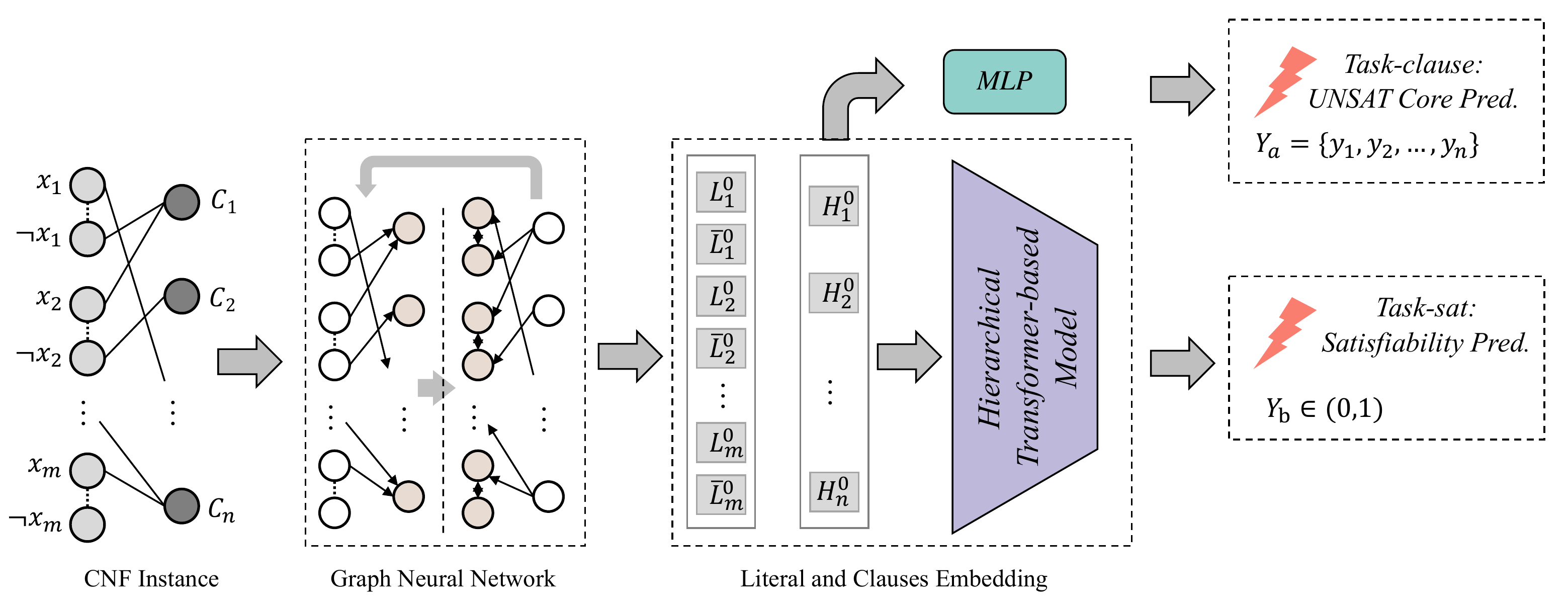}
	\caption{An overview of the model architecture.}
	\label{FIG:overview}
\end{figure*}
Fig.~\ref{FIG:overview} shows an overview of our proposed SATformer. 
Firstly, our model employs a GNN to obtain clause and literal representations from a CNF instance. After performing message-passing for several iterations based on the LCG, the GNN learns the literal embeddings with rich information from the connecting clauses and the clause embeddings containing information from the involved literals. 

% To determine whether a few clauses can make up an UNSAT core, the model shall learn a rich clause embedding containing the information from the involving literals. Consequently, we equip the NeuroSAT~\cite{selsam2018learning} as the GNN model, which performs message passing between clauses and literals for multiple iterations. It should be noticed that the GNN model in our SATformer only performs message pass for $10$ iterations instead of $26$ iterations as the original NeuroSAT. 

Secondly, after obtaining the clause embeddings, we design a classifier to predict which clauses are involved in the UNSAT core. Formally, the classifier is implemented by a Multilayer Perceptron (MLP) as shown in Eq.~\eqref{eq:coreCls}, where $H_i^0$ is the embedding of clause $C_i$ produced by the GNN, $y_i \in (0, 1)$ is the predicted probability that $C_i$ belongs to the UNSAT core, and $n$ is the number of clauses. It should be noted that this structure may not be perfect for classifying clauses with high accuracy. However, similar to NeuroCore~\cite{selsam2019guiding}, which learns a heuristic that assigns higher priority to important variables, the proposed classifier can help to highlight the clauses that contribute more to unsatisfiability. 

\begin{equation} \label{eq:coreCls}
y_i = \textit{MLP}(H_i^0), i=0, 1, ..., n
\end{equation}

Thirdly, we consider that the unsatisfiability arises from the interactions among clauses in the UNSAT core. We take the clause embeddings from the GNN as inputs and apply the Transformer to capture such clause relationships. Benefiting from the self-attention mechanism inside the Transformer, the SATformer gives more attention to the clauses that are likely to be involved in the UNSAT core, which is proved in Section~\ref{Sec:Exp:tf}. However, the standard Transformer only learns pairwise correlations, and there is almost no cause of unsatisfiability arising from a pair of clauses. To address this issue, we merge clauses into groups in a hierarchical manner and train a hierarchical Transformer-based model to learn these clause groups, which is elaborated in Section~\ref{Sec:Method:Hier}. 

\subsection{Training Objective} \label{Sec:Method:Train}
We adopt a multi-task learning (MTL)~\cite{caruana1997multitask} strategy to improve the performance of SATformer. As shown in Fig.~\ref{FIG:overview}, we train our model to predict UNSAT core (\textit{Task-clause}) and satisfiability (\textit{Task-sat}) simultaneously. 

Firstly, although the UNSAT core is not unique for a given instance, there are only a few possible combinations for minimal unsatisfiable cores (MUC), i.e., the UNSAT cores with the smallest size. We denote $n$ as the number of clauses and label a binary vector $Y_{clause}^{*} = \{y_1^{*}, y_2^{*}, ..., y_n^{*}\}$ as the ground truth for each instance, which represents whether the clause is included in MUC. Hence, the binary vector $Y_{clause}^{*}$ for a satisfiable instance consists of all \textit{zeroes} while some bits are assigned \textit{ones} for an unsatisfiable instance. 

We denote the clause embedding matrix as $\mathbf{H}^0 = \{ H_1^0, H_2^0, ..., H_n^0\}$ and obtain $Y_{clause} = \{y_1, y_2, ..., y_n\}$ to model a probability distribution over clauses using Eq.~\eqref{eq:task1}. We regard $Y_{clause}$ as the contribution of each clause to unsatisfiability, where a larger value indicates that the corresponding clause is more likely to form the MUC.

\begin{equation} \label{eq:task1}
Y_{clause} = \textit{softmax}(\textit{MLP}(\mathbf{H}^0))
\end{equation} 

The model is trained by minimizing the Kullback-Leibler (KL) divergence~\cite{kullback1951information}: 

\begin{equation} \label{eq:task1Loss}
\mathcal{L}_{clause} = \sum_{i=1}^{n} y_i^{*}\cdot \text{log}(\frac{y_i^{*}}{y_i})
\end{equation} 

Secondly, the SATformer predicts binary satisfiability as another task (\textit{Task-sat}). The model learns to produce a predicted result $Y_{sat} \in (0, 1)$ with single-bit supervision $Y_{sat}^{*} \in \{0, 1\}$. We apply Binary Cross-Entropy (BCE) loss to train the model. 
\begin{equation}
\mathcal{L}_{sat} = -(Y_{sat}^{*} \cdot \text{log} Y_{sat} + (1-Y_{sat}^{*}) \cdot \text{log} (1-(Y_{sat})))
\end{equation} 

Thus, the overall loss is the weighted sum of $\mathcal{L}_{clause}$ and $\mathcal{L}_{sat}$
\begin{equation} \label{eq:loss}
\mathcal{L} = \frac{p_{clause} \cdot \mathcal{L}_{clause} + p_{sat} \cdot \mathcal{L}_{sat}}{p_{clause} + p_{sat}}
\end{equation} 

Intuitively, \textit{Task-clause} aims to distinguish which clauses contribute more to unsatisfiability. By doing so, our model learns to pay more attention to these clauses and predicts the final satisfiability as \textit{Task-sat}. Therefore, we can improve the prediction accuracy by leveraging useful information between these two related tasks.

\subsection{Hierarchical Design} \label{Sec:Method:Hier}
We further propose a hierarchical Transformer-based model to capture the correlation among clauses. 
Formally, for the standard Transformer $\mathcal{T}$, we define the embeddings of the input tokens as $\mathbf{H} = \{H_1, H_2, ..., H_n\} \in \mathbb{R}^{n \times k}$. The self-attention matrix $\mathbf{A}$ and updated embeddings $\hat{\mathbf{H}} = (\hat{H_1}, \hat{H_2}, ..., \hat{H_n})$ are denoted as: 
\begin{equation}
    \begin{split}
        & \mathbf{Q} = \mathbf{W}^Q \mathbf{H},\ \mathbf{K} = \mathbf{W}^K \mathbf{H},\  \mathbf{V} = \mathbf{W}^V \mathbf{H} \\
        & \mathbf{A} = \textit{softmax}(\frac{\mathbf{Q} \mathbf{K}^\top}{\sqrt{k}}) \\
        & \hat{\mathbf{H}} = \mathcal{T}(\mathbf{H}) = \mathbf{A} \mathbf{V}
    \end{split}
\end{equation}
where $\mathbf{W}^Q$, $\mathbf{W}^K$, $\mathbf{W}^V$ are three learnable projection matrices. $\mathbf{Q}$, $\mathbf{K}$, $\mathbf{V}$ denote matrices packing sets of queries, keys, and values, respectively. And in our case, they have the same shape with $\mathbf{H}$.

\begin{figure}[!t]
	\centering
	\includegraphics[width=0.9\linewidth]{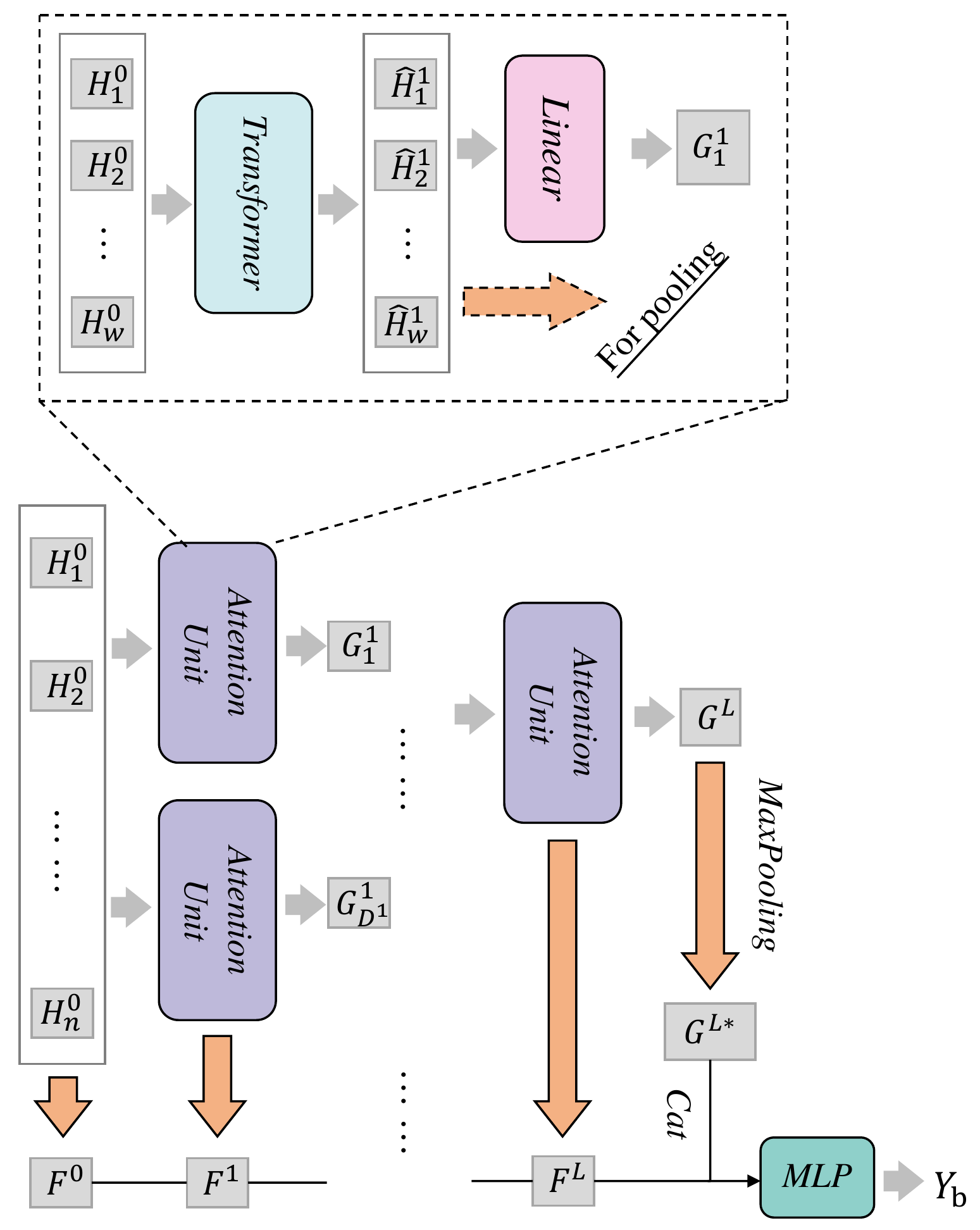}
	\caption{The architecture of the hierarchical Transformer-based model.}
	\label{FIG:hier}
\end{figure}

In most cases, the unsatisfiability can not be determined by a pair of clauses. As demonstrated in the example instance in Eq.~\eqref{eq:unsatcore}, the solver requires to consider at least $3$ clauses ($C_2$, $C_5$ and $C_6$) to assert unsatisfiable. However, capturing the dependencies among multiple tokens within one Transformer block is impractical. Although increasing the layer of Transformer blocks can mitigate this problem, it significantly increases the model complexity. In this paper, we propose a hierarchical Transformer-based model including many Attention Unit (AU), as shown in Fig.~\ref{FIG:hier}. Briefly, AU merges clauses within a fixed-size window into a clause group and then packages these groups into another group level by level. The resultant clause group embedding contains information from all clauses covered by the lower-level groups. At a higher level, the model takes clause group embeddings as inputs and learns the relationship between two clause groups. 

More formally, the clause embeddings from the GNN model are denoted as $H_1^0, ..., H_n^0$, where $n$ is the number of clauses. We also denote the clause group embeddings produced by AU as $G_1^l, ..., G_{D^l}^l$, where $D^l$ is the number of windows in level $l$ and the maximum level is $L$. Given the fixed window size $w$, we calculate $D^l$ as Eq.~\eqref{Eq:maxWinCnt}. Before dividing clauses into groups, we randomly pad the length of the embedding matrix to $w \cdot D^l$. For example, if there are $n=11$ clauses and the window size $w=4$, we divide these $11$ clauses with $1$ more padding clause into $D^1 = 3$ windows at the first level. 
\begin{equation} \label{Eq:maxWinCnt}
    D^l = \lceil \frac{n}{\textit{Pow(w, l)}} \rceil
\end{equation}

With a window of size $w$, the clause group embeddings are updated as: 
\begin{equation}
    \begin{split}
        \hat{\mathbf{H}}^{l}_d & = \mathcal{T}(\mathbf{H}^{l-1}_d) \\
        \mathbf{G}_d^{l} & = \textit{LN}(\hat{\mathbf{H}}^{l}_d)
    \end{split}
\end{equation}
where $d$ is the index, matrix $\mathbf{H}^{l-1}_d = \{H_{(d-1) \cdot w}^{l-1}, ..., H_{d\cdot w - 1}^{l-1}\}$ consists of embedding vectors of clauses in the window $d$ and the matrix $\hat{\mathbf{H}}^{l}_d$ after processed by Transformer block. We also combine these clause embeddings into one group embedding $\mathbf{G}_d^l$ for each window, which is implemented via a linear transformation \textit{LN}. For the higher level, i.e., $l \geq 2$, the inputs for the Transformer should be clause group embeddings $\mathbf{G}^{l-1}_d = \{G_{(d-1) \cdot w}^{l-1}, ..., G_{d\cdot w - 1}^{l-1}\}$ instead of $\mathbf{H}^{l-1}_d$. 

We still take the instance in Eq.~\eqref{eq:unsatcore} as an example. When the window size is $3$, these $9$ clauses are divided into $3$ clause groups, where $G_1^1$ contains the information from $\{C_1, C_2, C_3\}$, $G_2^1$ represents $\{C_4, C_5, C_6\}$, and so on. We cannot find any UNSAT cores by combining clauses pairwise, but at a high level, the clauses contained in $G_1^1$ and $G_2^1$ can construct an UNSAT core. Therefore, to determine whether the instance is satisfiable, the model should consider the pairwise relationships of groups at various grain sizes. 

As shown in Eq.~\eqref{Eq:levelembedding}, similar to the Feature Pyramid Networks (FPN)~\cite{lin2017feature}, we also obtain level embedding $F^l$ based on the maximum pooling (\textit{MaxPooling}) of all $\mathbf{H}^l$,  $\hat{\mathbf{H}}^l$ and $\hat{\mathbf{G}}^l$ as belows: 
\begin{equation} \label{Eq:levelembedding}
    F^l=\left\{
    \begin{array}{ll}
        \textit{MaxPooling} ({\mathbf{H}}^l_0, ..., {\mathbf{H}}^l_{D^l}), & l=0 \\
        \textit{MaxPooling} (\hat{\mathbf{H}}^l_0, ..., \hat{\mathbf{H}}^l_{D^l}), & l=1 \\
        \textit{MaxPooling} (\hat{\mathbf{G}}^l_0, ..., \hat{\mathbf{G}}^l_{D^l}), & l=2, ..., L 
    \end{array}
    \right.
\end{equation} 

Finally, a \textit{MLP} reads out the concatenation (\textit{Cat}) of $F^l$ ($l=1, 2, ..., L$) and the final clause group embedding $G^{L*}$. If there are multiple clause groups in the final level, we need to apply maximum pooling to obtain the final $G^{L*}$. 
\begin{equation} \label{Eq:readout}
    \begin{split}
        G^{L*} & = \textit{MaxPooling} (\mathbf{G}^L_0, ..., \mathbf{G}^L_{D^l}) \\
        Y_{sat} & = \textit{MLP} (\textit{Cat}(F^0, ..., F^L, G^{L*}))
    \end{split}
\end{equation} 

With the hierarchical Transformer-based model, our proposed SATformer can capture the causes of unsatisfiability among multiple clauses rather than pairs of clauses. 

%%%%%%%%%%%%%%%%%%%%%%%%%%%%%%%%%%%%%%%%%%%%%%%%%%%%%%%%%%%%%%%%%%%%%%%%%%%%%%%%%%%%%%
%%%%%%%%%%%%%%%%%%%%%%%%%%%%%%%%%%%%%%%%%%%%%%%%%%%%%%%%%%%%%%%%%%%%%%%%%%%%%%%%%%%%%%
\section{Combination with SAT solvers} \label{Sec:Comb}
In this subsection, we employ the SATformer as an initialization heuristic and combine it with non-learning SAT solvers. To facilitate understanding, we briefly introduce the background of modern non-learning SAT solvers and related heuristics in Section~\ref{Sec:Comb:Solver}. Following that, we demonstrate how to combine our SATformer with two moderns solvers (CaDiCal~\cite{queue2019cadical} and Kissat~\cite{fleury2020cadical}) in Section~\ref{Sec:Comb:Hybrid}

\subsection{Background on SAT Solvers} \label{Sec:Comb:Solver}
The mainstream modern SAT solvers~\cite{audemard2018glucose, sorensson2005minisat, queue2019cadical, fleury2020cadical} are based on the searching algorithm and reduce the search space by a heuristic named Conflict-Driven Clause Learning (CDCL)~\cite{marques2021conflict}. Specifically, CDCL-based SAT solvers first assign variables and perform Boolean constraint propagation until the current variable assignment is not valid, resulting in a \textit{conflict}. Then, the solvers analyze the conflict to identify variable dependencies and parse them into new clauses, referred to as \textit{learnt clauses}. Adding such learned clauses can introduce additional constraints into the solving procedure and significantly reduce the search space, especially for UNSAT problems~\cite{lu2003circuit}. 

In order to encounter conflicts as much as possible and speed up SAT solving, previous SAT solvers introduce a heuristic~\cite{moskewicz2001chaff} to guide variable decision for more conflicts. This approach maintains a Variable State-Independent Decaying Sum (VSIDS) score for each variable, indicating how many conflicts the variable has been involved in. During the solving procedure, the solvers assign decision values to variables with high VSIDS scores as a priority. Due to the promising performance of this variable decision heuristic, many modern SAT solvers also use VSIDS scores as a variable decision heuristic~\cite{queue2019cadical, fleury2020cadical, sorensson2005minisat}. However, the VSIDS scores are only obtained by an online algorithm during solving and lack an efficient initialization strategy. 

\subsection{Hybrid Solver Design} \label{Sec:Comb:Hybrid}
We combine our SATformer with modern SAT solvers. Firstly, our SATformer predicts both binary satisfiability result and the clauses' contribution to unsatisfiability. As the solving procedure for UNSAT instances is to cause more conflicts to reduce the search space, the contribution to unsatisfiability (see Eq.~\eqref{eq:task1}) can be considered as the contribution to conflicts.

Secondly, we convert the above clause prediction to variable prediction. Specifically, based on the clause prediction, we calculate the score leading to conflicts for each variable using Eq.~\eqref{Eq:VSIDS}, where $A_{j, i}$ is the adjacency matrix of variable $x_j$ and clause $C_i$, and $M$ is the total number of variables. 
\begin{equation} \label{Eq:VSIDS}
    v_j = \sum_{C_i \in \phi}(A_{j, i} * y_i), j = 1, ..., M
\end{equation}

Thirdly, we provide the above variable score $v_j$ as the initial VSIDS score to the modern SAT solvers. Since most solvers reserve the interface of VSIDS score, our SATformer can be embedded as a plug-in without modifying the SAT solvers. The details of the combination are shown in Algorithm.~\ref{Algo:Solver}.

\begin{algorithm}[htbp]
    \caption{VSIDS Initialization with SATformer}
    \label{Algo:Solver}
    Problem instance $\phi$, with $M$ variables and $n$ clauses. \\
    Variable scores $v_j, j=1,\ldots,M$, initialized with zeroes. \\
    SAT solver $\mathcal{S}$. 
    \begin{algorithmic}[1]
    \STATE $Y_{sat}, Y_{clause} = \text{SATformer}(\phi)$
    \IF{$Y_{sat} == \text{UNSAT}$}
        \FOR{$j = 1 \rightarrow M$}
            \FOR{$i = 1 \rightarrow n$}
                \IF{Variable $x_j$ is connected with Clause $C_i$}
                    \STATE $v_{j} = v_{j} + y_{i}$ 
                \ENDIF
            \ENDFOR
        \ENDFOR
    \ENDIF
    \STATE \ 
    \STATE Initialize solver $\mathcal{S}$ with instance $\phi$: $\mathcal{S}.input(\phi)$
    \STATE Update the initial VSIDS in $\mathcal{S}$ with $v$: $\mathcal{S}.updateScore(v)$
    \STATE Perform solving of $\mathcal{S}$: $\mathcal{S}.run()$
    \STATE Return results: $\mathcal{S}.results$
    \end{algorithmic}
\end{algorithm}

\section{Experiments} \label{Sec:Exp}
In this section, we conduct three experiments in three parts, with detailed experimental settings provided in Section~\ref{Sec:Exp:setting}. Firstly, we demonstrate the ability of SATformer to act as a standalone satisfiability classifier by comparing it with another end-to-end SAT solver,  NeuroSAT~\cite{selsam2018learning} (see Section~\ref{Sec:Exp:compare}). Secondly, we investigate the effectiveness of our model design through a series of ablation studies (see Section~\ref{Sec:Exp:tf}, \ref{Sec:Exp:Hier} and \ref{Sec:Exp:MuTask}). Thirdly, we integrate SATformer with modern SAT solvers: CaDiCaL~\cite{queue2019cadical} and Kissat~\cite{fleury2020cadical}, which are the state-of-the-art solvers to the best of our knowledge. As shown in Section~\ref{Sec:Exp:Solver}, the hybrid SAT solvers are employed to prove the unsatisfability of hard logic equivalence checking (LEC) instances. Section~\ref{Sec:Exp:Limit} discusses our current limitations and proposes potential solutions for future work.

\subsection{Experimental setting} \label{Sec:Exp:setting}
We present the details of the experimental setting, including the construction of datasets and the hyperparameters of the SATformer model. In the following experiments, we adopt the same settings described here for all models unless otherwise specified.  

\textbf{Evaluation metric}: Our SATformer predicts the satisfiability of a given instance. We construct both satisfiable and unsatisfiable instances in the testing dataset and record the accuracy of binary classification. 

\textbf{Dataset preparation}: We generate $10K$ satisfiable instances and $10K$ unsatisfiable instances. Following the same dataset generation scheme as~\cite{selsam2018learning}, a pair of random $k$-SAT satisfiable and unsatisfiable instances is generated together with only one different edge connection. Here, SR($m$) indicates that the instance contains $m$ variables. In our training dataset, we set the problem size as SR($3$-$10$). Furthermore, we enumerate all possible clause subsets of each instance and label the minimal UNSAT cores. We also generate default testing datasets, each containing $50$ satisfiable and $50$ unsatisfiable instances in SR($3$-$10$), SR($20$), SR($40$) and SR($60$), respectively. 

\textbf{Implementation Details}: In the GNN structure, we adopt the same configurations as NeuroSAT~\cite{selsam2018learning}, except for reducing the number of message-passing iterations from $26$ to $10$. In the Hierarchical Transformer structure, we directly use clause embeddings as input tokens, resulting in the same hidden state dimension of $128$. The window size $w$ is set to $4$, and the total number of levels in the hierarchical structure is $4$. Inside each Transformer block, we set the number of heads in the multi-head self-attention mechanism to $8$. The MLPs used to produce the two predictions for \textit{Task-clause} and \textit{Task-sat} are both configured as 3 layers. We train the model for $80$ epochs with a batch size $16$ on $4$ Nvidia V100 GPUs. We adopt the Adam optimizer~\cite{kingma2014adam} with a learning rate $10^{-4}$ and weight decay $10^{-10}$. 

\textbf{Model Complexity}: The complexity of NeuroSAT and above SATformer are illustrated in Table~\ref{Tab:FLOPs}, where \# FLOPs is the number of floating-point operations and \# Para. is the total number of trainable parameters. Although our model requires more trainable parameters to build up Transformer structures, SATformer performs faster (lower \# FLOPs) than NeuroSAT due to the fewer message-passing iterations. 

\begin{table}[!t]
\centering
\caption{The Model Complexity of NeuroSAT and SATformer} \label{Tab:FLOPs}
\begin{tabular}{@{}llll@{}}
\toprule
           & NeuroSAT & SATformer  \\ \midrule
\# Param.  & 429.31 K  & 732.47 K   \\ 
\# FLOPs  & 207.91 M & 152.93 M    \\\bottomrule
\end{tabular}
\end{table}

\subsection{Performance Comparison with NeuroSAT} \label{Sec:Exp:compare}
We compare the performance of SATformer and NeuroSAT~\cite{selsam2018learning} on satisfiability classification. We do not consider other standalone SAT solvers~\cite{amizadeh2018learning, ozolins2021goal, li2022deepsat} because they cannot produce only a single binary result. 

Both NeuroSAT and our SATformer require the input instances to be in CNF. While SR($m$) represents the \textit{problem scale}, we use another metric, $CV = \frac{n}{m}$, to better quantify the \textit{problem difficulty}, where $n$ is the number of clauses and $m$ is the number of variables. A higher $CV$ value indicates more clause constraints in the formula, making the instance more difficult to solve. The instances from the training dataset and the default testing dataset have $CV>5$. These satisfiable instances only have $1$ or $2$ possible satisfying assignments in total. Besides, we also produce $50$ satisfiable and $50$ unsatisfiable simplified instances with $CV=3$ and $CV=4$ for each problem scale. To ensure fairness, we train NeuroSAT on the same training dataset in SR($3$-$10$). The results of the binary classification are listed in Table~\ref{TAB:compNeurosat}.

\begin{table}[!t]
\renewcommand\tabcolsep{2.5pt}
\centering
\caption{Performance Comparison of NeuroSAT and SATformer} \label{TAB:compNeurosat}
\begin{tabular}{@{}llllll@{}}
\toprule
\textit{}           &                & SR($3$-$10$) & SR($20$) & SR($40$) & SR($60$)        \\ \midrule
$CV$\textgreater{}$5$ & NeuroSAT       & 87\%          & 61\%          & 58\%          & 50\%          \\
                    & SATformer      & 94\%          & 77\%          & 68\%          & 61\%          \\
                    & \textbf{Impr.} & \textbf{7\%}  & \textbf{16\%} & \textbf{10\%}  & \textbf{11\%}  \\ \midrule
$CV=4$               & NeuroSAT       & 83\%          & 59\%          & 58\%          & 50\%          \\
                    & SATformer      & 99\%          & 98\%          & 91\%          & 86\%          \\
                    & \textbf{Impr.} & \textbf{16\%} & \textbf{39\%} & \textbf{33\%} & \textbf{36\%} \\ \midrule
$CV=3$               & NeuroSAT       & 89\%          & 50\%          & 57\%          & 50\%          \\
                    & SATformer      & 99\%          & 98\%          & 98\%          & 98\%          \\
                    & \textbf{Impr.} & \textbf{10\%} & \textbf{48\%} & \textbf{98\%} & \textbf{48\%}  \\ \bottomrule
\end{tabular}
\end{table}

From Table~\ref{TAB:compNeurosat}, we have a few observations. Firstly, our SATformer achieves higher performance than NeuroSAT across all datasets. For example, while NeuroSAT can only correctly classify $67\%$ in SR($20$) and $56\%$ in SR($40$) when $CV>5$, our SATformer has an accuracy of $70\%$ in SR($20$) and $61\%$ in SR($40$). Secondly, the results of NeuroSAT have no significant difference when the problem difficulty is reduced, but our SATformer performs better on the simplified instances. For example, SATformer solves all instances with $CV=3$ regardless of the problem scale. The reason for this is that SATformer relies on the correlation among clauses to solve SAT problems. With fewer clauses, SATformer harnesses such correlations more easily. In contrast, NeuroSAT only learns the instance-level features with single-bit supervision and does not obtain richer representations with the reduction of problem difficulty. 

To conclude, our SATformer outperforms the NeuroSAT. Additionally, our SATformer shows a similar property to traditional SAT solvers, i.e., it can achieve higher performance on instances with lower difficulty. 

\subsection{Effectiveness of Transformer Blocks} \label{Sec:Exp:tf}
This section investigates the effectiveness of the Transformer Blocks in SATformer. On the one hand, we compare our SATformer with a derived model framework called \textit{SATformer-MLP}, which replaces Transformer blocks with MLPs of the equivalent number of model paremeters (\# Param.). Table~\ref{TAB:compMlp} shows the experimental results. Since the SATformer-MLP treats all clauses equally instead of selectively enhancing the correlation between some clauses, it shows an inferior performance compared to the original SATformer. 

\begin{table}[!t]
\centering
\caption{Performance Comparison of SATformer and SATformer-MLP} \label{TAB:compMlp}
\begin{tabular}{@{}lllll@{}} 
\toprule
\textit{} & SR($3$-$10$) & SR($20$) & SR($40$) & SR($60$) \\ \midrule
\textbf{SATformer} & \textbf{94\%}     & \textbf{77\%}   & \textbf{68\%}   & \textbf{61\%}   \\ 
SATformer-MLP & 90\%     & 72\%   & 60\%   & 52\%   \\ \bottomrule
\end{tabular}
\end{table}

On the other hand, as the Transformer distinctively treats token pairs based on the self-attention mechanism, we explore the attention weights assigned to different token pairs. 
Based on the \text{Task-clause} UNSAT core prediction, we equally divide the clauses into two categories: likely to form an UNSAT \underline{C}ore (denoted as \textit{C}-clause) and \underline{U}likely to form an UNSAT core (denoted as \textit{U}-clause) cause unsatisfiable. Therefore, the pairwise connections are classified into $4$ types: between \textit{C}-clause and \textit{C}-clause (\textit{CC}), \textit{C}-clause and \textit{U}-clause (\textit{CU}), \textit{U}-clause and \textit{C}-clause (\textit{UC}), \textit{U}-clause and \textit{U}-clause (\textit{UU}). 
We calculate the overall attention weights for these $4$ types, respectively, and list the percentages in Table~\ref{TAB:attn}. A higher percentage indicates that the model pays more attention to the corresponding pairwise correlation. The Transformer assigns almost $70\%$ of its attention weights to the \textit{CC} connection. Moreover, along with the \textit{CU} and \textit{UC} connections, the model allocates $97.23\%$ of its attention weights to capture the correlation related to \textit{C}-clauses. Therefore, our Transformer-based model mainly focuses on the clauses that contribute more to unsatisfiability.

\begin{table}[!t]
\centering
\caption{The Attention Weight Percentages of Four Type Connections} \label{TAB:attn}
\begin{tabular}{@{}llll@{}}
\toprule
\textit{CC} & \textit{CU} & \textit{UC} & \textit{UU} \\ \midrule
68.27\%     & 14.95\%     & 14.01\%     & 2.77\%      \\ \bottomrule
\end{tabular}
\end{table}

To summarize, our SATformer not only captures the relationship among clauses but also learns how to pay more attention to those clauses that are more likely to raise unsatisfiability.

\subsection{Effectiveness of Hierarchical Transformer} \label{Sec:Exp:Hier}
In this subsection, we conduct an ablation study and create some special cases to demonstrate the effectiveness of our Hierarchical Transformer. 

Firstly, we replace the hierarchical structure with plain Transformer blocks. This model takes clause embeddings as input tokens and updates them with $4$ (same as $L=4$ in default SATformer configuration) stacked Transformer blocks (denoted as \textit{SATformer-NoHier}). Then, the model produces an embedding vector by pooling all updated clause embeddings and predicts the final binary result. As shown in Table~\ref{TAB:compNoHier}, SATformer-NoHier can handle fewer instances than SATformer. For example, SATformer-NoHier only achieves $70\%$ accuracy in SR($3$-$10$) and performs as a randomly guessing classifier in SR($40$) and SR($60$) instances. On the contrary, SATformer is still effective in SR($40$) and SR($60$). The reason for this is that SATformer-NoHier can only learn pairwise dependencies, while a pair of clauses may not have a decisive effect on the satisfiability. Although the model can learn multiple relationships by stacking Transformer blocks, $4$ layers are still far from sufficient. 

\begin{table}[!t]
\renewcommand\tabcolsep{2.5pt}
\centering
\caption{Performance Comparison of SATformer and SATformer-NoHier} \label{TAB:compNoHier}
\begin{tabular}{@{}lllll@{}} 
\toprule
\textit{} & SR($3$-$10$) & SR($20$) & SR($40$) & SR($60$) \\ \midrule
\textbf{SATformer} & \textbf{94\%}     & \textbf{77\%}   & \textbf{68\%}   & \textbf{61\%}    \\ 
SATformer-NoHier & 70\%     & 57\%   & 50\%   & 50\%   \\ \bottomrule
\end{tabular}
\end{table}

Secondly, to further demonstrate the ability of SATformer to learn multiple relationships, we construct two equivalent unsatisfiable instances as Eq.~\eqref{eq:twocase} based on Eq.~\eqref{eq:unsatcore} with different clause orders. The three clauses $C_2$, $C_5$, and $C_6$ forming MUC are highlighted with underlines.

\begin{equation} \label{eq:twocase}
\footnotesize
  \begin{split}
\phi_{U1} & := C_1 \land \underline{C_2} \land C_3 \land C_4 \land \underline{C_5} \land \underline{C_6} \land C_7 \land C_8 \land C_9 \\
\phi_{U2} & := C_1 \land \underline{C_2} \land C_3 \land C_4 \land \underline{C_5} \land C_9 \land C_7 \land C_8 \land \underline{C_6}
  \end{split}
\end{equation}

\begin{figure}[!t]
	\centering
	\includegraphics[width=0.8\linewidth]{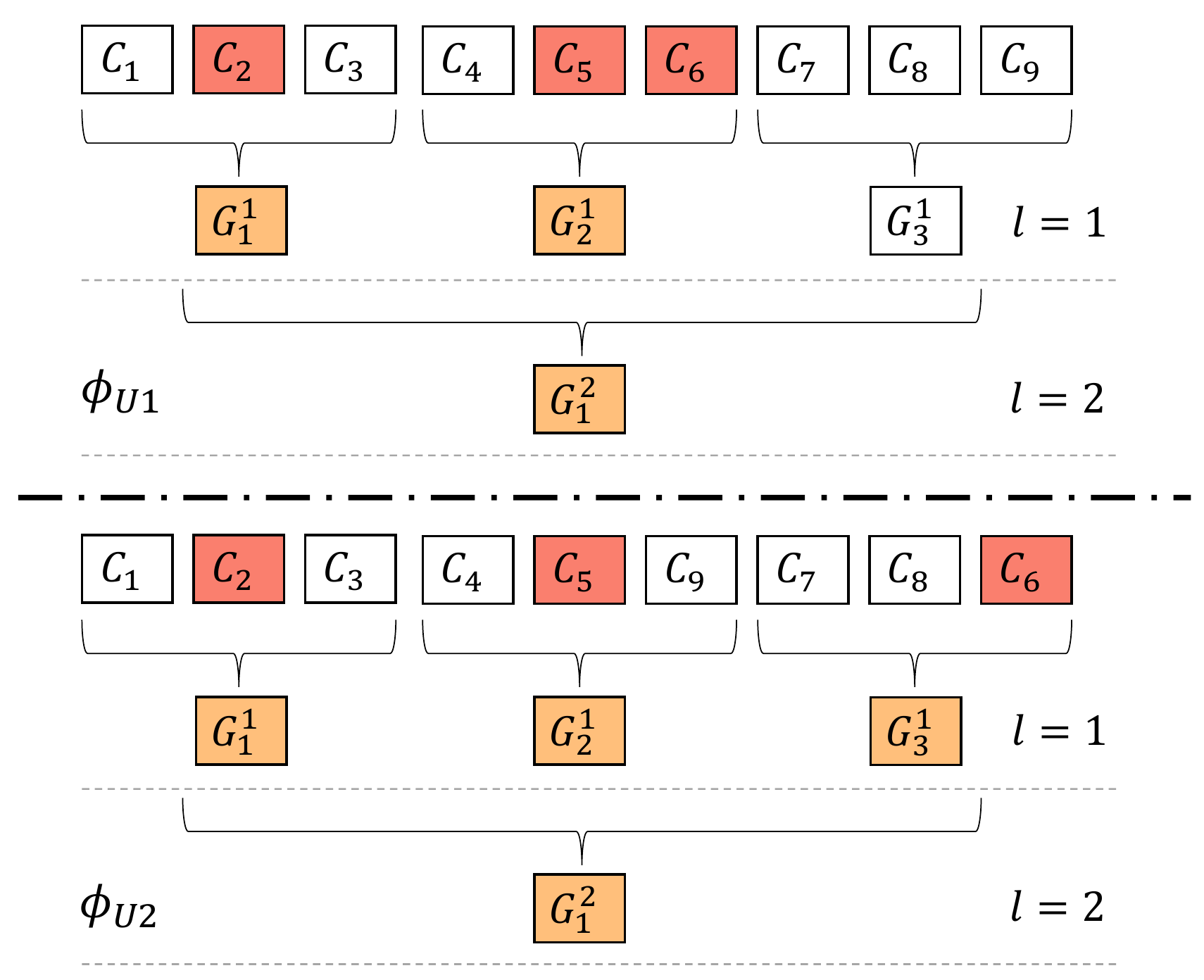}
	\caption{Two special unsatisfiable cases in hierarchical structure}
	\label{FIG:case}
    \vspace{-10pt}
\end{figure}

As shown in Fig.~\ref{FIG:case}, for the instance $\phi_{U1}$, the clauses (from $C_1$ to $C_6$) in the first two groups can construct an unsatisfiable sub-instance. Hence, by modeling the pairwise relationship between input token embeddings $G_1^1$ and $G^1_2$ in level $l=1$, the model can determine the satisfiability. However, there is no pairwise correlation leading to unsatisfiability in the first hierarchical level in $\phi_{U2}$. We train two additional SATformer models with a window size of $w=3$, but with different hierarchical levels: one with $L=1$ and the other with $L=2$. According to the experimental results, the model with $L=1$ which relies on the group embeddings in level $l=1$ can only correctly predict $\phi_{U1}$ as unsatisfiable. However, if we add one more Transformer block in level $l=2$, the model (with $L=2$) can learn the group embedding $G_1^2$, which contains information from these $3$ clause groups. In this case, the SATformer model can correctly predict both instances as unsatisfiable. 

In summary, our SATformer can learn the relationships among multiple tokens due to the pairwise self-attention mechanism and hierarchical structure. Besides, to mitigate the effect of clause ordering, we shuffle the clauses every $10$ training iterations.

\subsection{Effectiveness of Multi-task Learning Strategy} \label{Sec:Exp:MuTask}
To investigate the effectiveness of the MTL strategy in our SATformer, we train another model without UNSAT core supervision (denote as \textit{SATformer-NoCore}) and compare it with the original SATformer. Table~\ref{TAB:compNoCore} shows the experimental results. After removing the \textit{Task-clause} UNSAT core prediction, the model's performance on all four datasets decreases. 

\begin{table}[!t]
\renewcommand\tabcolsep{2.5pt}
\centering
\caption{Performance Comparison of SATformer and SATformer-NoCore} \label{TAB:compNoCore}
\begin{tabular}{@{}lllll@{}} 
\toprule
\textit{} & SR($3$-$10$) & SR($20$) & SR($40$) & SR($60$) \\ \midrule
\textbf{SATformer} & \textbf{94\%}     & \textbf{77\%}   & \textbf{68\%}   & \textbf{61\%}    \\ 
SATformer-NoCore & 89\%     & 67\%   & 57\%   & 51\%   \\ \bottomrule
\end{tabular}
\end{table}

% Generally, as we demonstrate the self-attention mechanism with attention weight in Sec.~\ref{Sec:Exp:tf}, predicting the clause contribution to unsatisfiability is significantly related to SAT solving. Therefore, the MTL strategy benefits the model performance.

Generally, UNSAT core prediction retains rich information about the contribution of clauses to unsatisfiability, and it is significantly related to SAT solving. Therefore, the MTL strategy incorporating UNSAT core prediction benefits the model's performance.

%%%%%%%%%%%%%%%%%%%%%%%%%%%%%%%%%%%%%%%%%%%%%%%%%%%%%%%%%%%%%%%%%%%%%%%%%%%%%%%%%%%%%%
%%%%%%%%%%%%%%%%%%%%%%%%%%%%%%%%%%%%%%%%%%%%%%%%%%%%%%%%%%%%%%%%%%%%%%%%%%%%%%%%%%%%%%

\subsection{Combination with SAT Solvers} \label{Sec:Exp:Solver}
\begin{table*}[!t]
\centering
\caption{Average Runtime Comparison between the Solvers w/o and w/ SATformer} \label{TAB:Exp:Solver}
\vspace{-5pt}
\begin{tabular}{@{}lll|llll|llll@{}}
\toprule
             &                 &               & \multicolumn{4}{c|}{CaDiCaL}                                           & \multicolumn{4}{c}{Kissat}                                            \\
             & Size            & Model(s)      & w/o (s)     & w/ (s)        & Overall (s)      & Reduction        & w/o (s)     & w/ (s)        & Overall (s)      & Reduction       \\ \midrule
UNSAT        & 20,114          & 2.30          & 967.00          & 755.43          & 757.72          & 21.64\%          & 787.73          & 723.67          & 725.97          & 7.84\%          \\
SAT          & 20,102          & 2.05          & 6.63            & 6.62            & 8.67            & -30.77\%         & 3.79            & 3.78            & 5.83            & -53.77\%       \\
\textbf{All} & \textbf{20,112} & \textbf{1.94} & \textbf{486.81} & \textbf{381.02} & \textbf{382.96} & \textbf{21.33\%} & \textbf{395.76} & \textbf{363.73} & \textbf{365.67} & \textbf{7.60\%} \\ \bottomrule
\end{tabular}
\vspace{-5pt}
\end{table*}

\begin{table*}[!ht]
\centering
\caption{Searching Process Comparison between the CaDiCaL w/o and w/ SATformer} \label{TAB:Exp:Failcase}
\vspace{-5pt}
\begin{tabular}{@{}l|lll|lll|ll@{}}
\toprule
          & \multicolumn{3}{c|}{w/o SATformer} & \multicolumn{3}{c|}{w/ SATformer} & \# Lemma  & Time      \\
Instances & Core Size  & \# Lemma  & Time (s)  & Core Size  & \# Lemma  & Time (s) & Reduction & Reduction \\ \midrule
C1        & 1343       & 601331    & 18.91     & 1341       & 557498    & 17.01    & 7.29\%    & 10.06\%   \\
C2        & 2755       & 935965    & 28.13     & 2753       & 663373    & 23.05    & 29.12\%   & 18.07\%   \\ \midrule
C3        & 1836       & 677794    & 20.03     & 1834       & 782211    & 21.75    & -15.41\%  & -8.59\%   \\
C4        & 2744       & 717451    & 21.33     & 2739       & 852268    & 23.28    & -18.79\%  & -9.12\%   \\ \bottomrule
\end{tabular}
\vspace{-10pt}
\end{table*}

In this subsection, we combine SATformer with CaDiCaL~\cite{queue2019cadical} and Kissat~\cite{fleury2020cadical} to demonstrate its effectiveness. The solvers integrated with SATformer are denoted as \textit{w/ SATformer}, while the baseline solvers without SATformer are denoted as \textit{w/o SATformer}. To ensure fairness, the runtime of the w/ SATformer solver includes the model inference time. 
We test these solvers with $60$ industrial LEC instances, comprising of $30$ SAT instances and $30$ UNSAT instances. 

\begin{figure}
    \centering
    \includegraphics[width=0.9\linewidth]{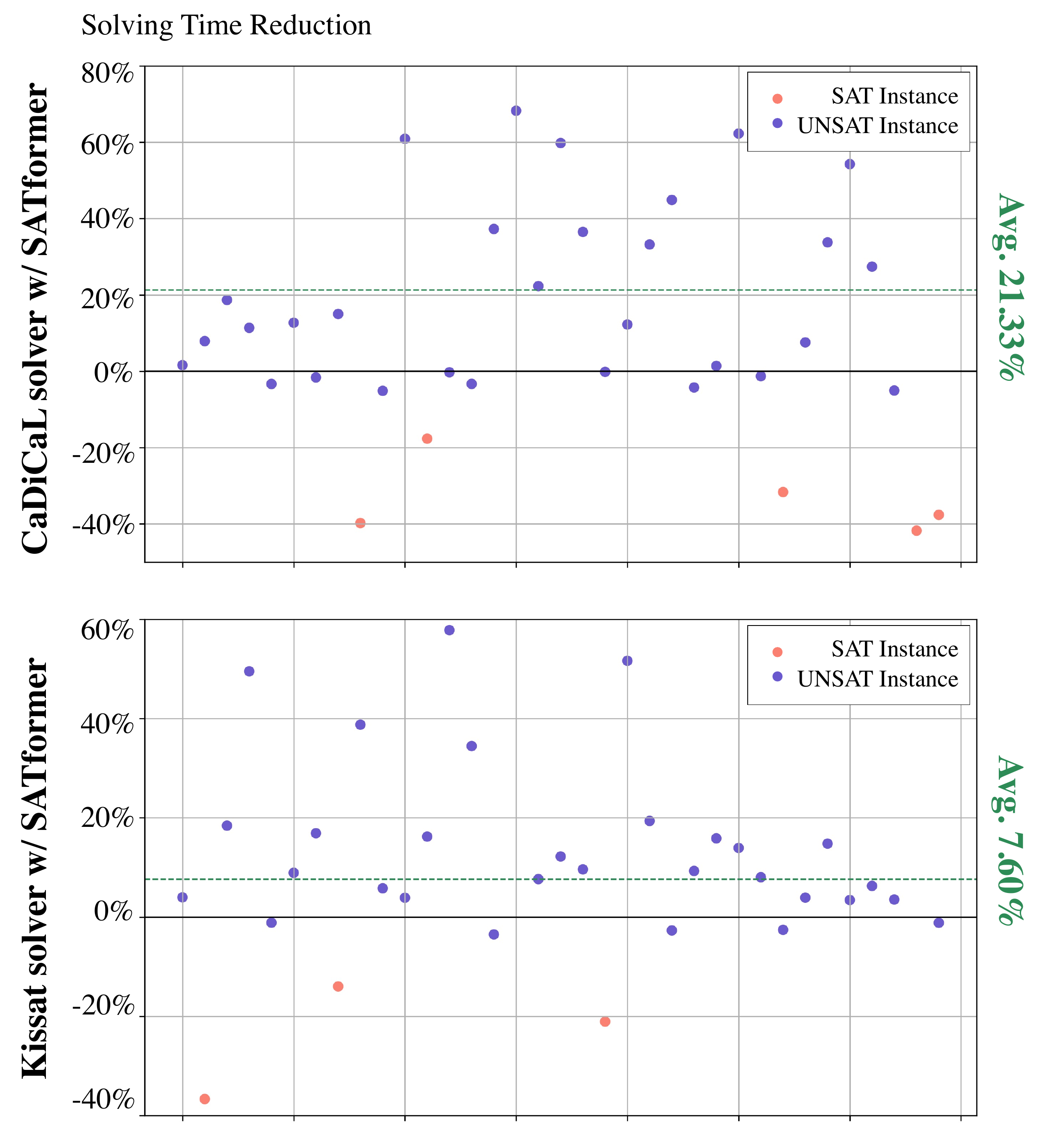}
    \caption{The runtime reduction of SATformer on CaDiCal~\cite{queue2019cadical} (upper) and Kissat~\cite{fleury2020cadical} (below)} 
    \label{FIG:EXP:Cadical}
    \vspace{-10pt}
\end{figure}

Fig.~\ref{FIG:EXP:Cadical} shows the runtime reduction achieved by SATformer on different solvers, where the red points represent SAT instances and the blue points represent UNSAT instances. To ensure better visualization, we ignore some cases with extremely large runtime increase or decrease. Both CaDiCaL and Kissat can achieve runtime reduction by integrating SATformer in most cases. However, there are still a few cases (especially in satisfiable instances) indicating a performance degradation when combining the solvers with SATformer. 

% \begin{figure}[!t]
%     \centering
%     \includegraphics[width=0.9\linewidth]{figs/cadical.pdf}
%     \caption{The runtime reduction of SATformer on CaDiCaL~\cite{queue2019cadical}}
%     \label{FIG:EXP:Cadical}
% \end{figure}
% \begin{figure}[!ht]
%     \centering
%     \includegraphics[width=0.9\linewidth]{figs/kissat.pdf}
%     \caption{The runtime reduction of SATformer on Kissat~\cite{fleury2020cadical}}
%     \label{FIG:EXP:Kissat}
% \end{figure}

To further elucidate the above phenomenon and evaluate the effectiveness of our SATformer, we calculate the average runtime for UNSAT instances, SAT instances and all instances, respectively, as shown in Table~\ref{TAB:Exp:Solver}. The table reveals several key observations. 
Firstly, compared to solvers w/o SATformer, CaDiCaL and Kissat w/ SATformer achieve a speedup of $21.64\%$ and $7.60\%$, respectively, when solving UNSAT instances. Therefore, SATformer can improve the efficiency of UNSAT solving. 
Secondly, SATformer results in a runtime increase for both CaDiCaL and Kissat solvers. Taking CaDiCaL as an example, although w/ SATformer requires $30.77\%$ more runtime than w/o SATformer, the initialization heuristic does not have a negative impact on the solver. The gap in solving time between w/ SATformer ($6.62$s on average) and w/o SATformer ($6.63$s on average) is quite small. Since the satisfiable instances in EDA are naturally easy to solve, the efficiency improvement brought by the heuristic is not significant.
Thirdly, the inference time of SATformer accounts for only a small fraction of the overall solving time, with $0.51\%$ in CaDiCaL and $0.53\%$ in Kissat, respectively. Moreover, SATformer is able to infer within polynomial time that is only proportional to the size of instances. Thus, the overhead of SATformer is within an acceptable range. 
In summary, our SATformer can significantly improve the efficiency of UNSAT solving  without harming SAT solving.

\subsection{Failure Case Analysis and Discussion} \label{Sec:Exp:Limit}
% 1. guide得到的core不一样，不同的搜索路径
% 2. mult效果不好，是XOR多

In this subsection, we analyze the limitations of the current SATformer and propose potential solutions to further improve SATformer in the future. 
On the one hand, as shown in Table~\ref{TAB:Exp:Solver}, SATformer cannot improve the efficiency of solving satisfiable problems. This is because the heuristic objective of finding a satisfying assignment is different from the objective of proving unsatisfiability~\cite{oh2015between}, and the prediction of the UNSAT core is meaningless for satisfiable instances. To improve the generalization ability of SATformer, we can design a heuristic based on decoding assignments~\cite{selsam2018learning}, especially for satisfiable instances. As a result, our next version of SATformer can predict satisfiability and accelerate the solving process for both SAT and UNSAT instances, allowing us to select the appropriate heuristic based on the satisfiability of the instance.

On the other hand, as shown in Fig.~\ref{FIG:EXP:Cadical}, we can find that integrating SATformer may reduce the efficiency of solving UNSAT instances in several cases. We investigate the reason for these failure cases with four representative instances.  
% 我们选择4个有代表性的instances，并列出使用不同的CaDiCaL求解器，求解过程所需时间和生成lemma的数量。每一条lemma由搜索中的冲突分析生成，lemma的数量用于描述求解中陷入conflict的次数，也就是搜索路径中回溯的次数。根据表1的结果，我们发现在C1和C2上，SATformer能够指导出一条寻找更小UNSAT Core的搜索路径，并减少lemma数量，降低搜索时间。而在C3和C4上，即使SATformer找到了更小的UNSAT core，但搜索需要更多次回溯，也就消耗了更多的时间。这是因为搜索时间与UNSAT core的大小正相关，而非绝对正比于UNSAT core大小~\cite{nadel2010boosting}。因此SATformer会改变原本的搜索路径，可能会导致derive a small but hard-to-prove UNSAT core。
Table~\ref{TAB:Exp:Failcase} presents the size of the UNSAT core (Core Size), solving time (Time) and the number of generated lemmas (\# Lemma) using CaDiCaL solvers w/o and w/ SATformer. Each lemma is generated by conflict analysis in the search process. The number of lemmas has a positive correlation with the number of conflicts encountered during solving and the number of times the search backtracks. According to the results in Table~\ref{TAB:Exp:Failcase}, we observe that for C1 and C2, SATformer guides the search towards a smaller UNSAT core and reduce the number of lemmas by $7.29\%$ and $29.12\%$, respectively, leading to a runtime reduction of $10.06\%$ and $18.07\%$, respectively. However, for C3 and C4, although SATformer tends to find a smaller UNSAT core, more backtracking is required during the search, which consumes more runtime. Since the search time is positively correlated with the size of the UNSAT core, rather than being absolutely proportional to the size of the UNSAT core~\cite{nadel2010boosting}, SATformer could derive a small but hard-to-prove UNSAT core. Consequently, to further improve the performance of SATformer in the future, the SATformer should guide the search path towards the easiest UNSAT core rather than the minimal UNSAT core.

\section{Conclusion} \label{Sec:Conclusion}
% This paper presents \textit{SATformer}, a Transformer-based framework for SAT problems. We focus on UNSAT core learning, i.e., identifying the existence of an unsatisfiable core via the modeling of clause interactions. Specifically, SATformer effectively tokenizes each clause into embeddings through a GNN and captures clause correlations utilizing a hierarchical Transformer-based model. By leveraging both single-bit satisfiability and the unsatisfiable core as supervisory elements, SATformer can effectively predict the satisfiability of an instance and quantify each clause's contribution to unsatisfiability. SATformer outperforms NeuroSAT as an end-to-end learning-based satisfiability classifier. Furthermore, we integrate SATformer as an initialization heuristic into modern heuristic-based SAT solvers. Our experimental results show that SATformer can reduce the solving time by an average of $21.33\%$ for the CaDiCaL solver and $7.60\%$ for the Kissat solver on the logic equivalence checking task.

This paper proposes \textit{SATformer}, a novel Transformer-based framework for predicting satisfiability and accelerating the solving process for UNSAT problems. The focus of our work is on UNSAT core learning, which involves identifying the existence of an unsatisfiable core by modeling clause interactions. Specifically, SATformer maps clauses into embeddings through a graph neural network (GNN) and captures clause correlations using a hierarchical Transformer-based model. By leveraging both single-bit satisfiability and the unsatisfiable core as supervisions, SATformer can effectively predict the satisfiability of an instance and quantify each clause's contribution to unsatisfiability
Our results demonstrate that SATformer outperforms NeuroSAT as an end-to-end learning-based satisfiability classifier. Furthermore, we integrate SATformer as an initialization heuristic into modern SAT solvers. Our experimental results show that SATformer can reduce the solving time by an average of $21.33\%$ for the CaDiCaL solver and $7.60\%$ for the Kissat solver on the logic equivalence checking task.

\clearpage
\balance

\section*{Acknowledgments}
This work was supported in part by the General Research Fund of the Hong Kong Research Grants Council (RGC) under Grant No. 14212422 and in part by Research Matching Grant CSE-7-2022.

\bibliographystyle{IEEEtran}
\bibliography{ref}

% Generated by IEEEtran.bst, version: 1.14 (2015/08/26)
\begin{thebibliography}{10}
\providecommand{\url}[1]{#1}
\csname url@samestyle\endcsname
\providecommand{\newblock}{\relax}
\providecommand{\bibinfo}[2]{#2}
\providecommand{\BIBentrySTDinterwordspacing}{\spaceskip=0pt\relax}
\providecommand{\BIBentryALTinterwordstretchfactor}{4}
\providecommand{\BIBentryALTinterwordspacing}{\spaceskip=\fontdimen2\font plus
\BIBentryALTinterwordstretchfactor\fontdimen3\font minus
  \fontdimen4\font\relax}
\providecommand{\BIBforeignlanguage}[2]{{%
\expandafter\ifx\csname l@#1\endcsname\relax
\typeout{** WARNING: IEEEtran.bst: No hyphenation pattern has been}%
\typeout{** loaded for the language `#1'. Using the pattern for}%
\typeout{** the default language instead.}%
\else
\language=\csname l@#1\endcsname
\fi
#2}}
\providecommand{\BIBdecl}{\relax}
\BIBdecl

\bibitem{goldberg2001using}
E.~I. Goldberg, M.~R. Prasad, and R.~K. Brayton, ``Using sat for combinational
  equivalence checking,'' in \emph{Proceedings Design, Automation and Test in
  Europe. Conference and Exhibition 2001}.\hskip 1em plus 0.5em minus
  0.4em\relax IEEE, 2001, pp. 114--121.

\bibitem{mcmillan2003interpolation}
K.~L. McMillan, ``Interpolation and sat-based model checking,'' in
  \emph{Computer Aided Verification: 15th International Conference, CAV 2003,
  Boulder, CO, USA, July 8-12, 2003. Proceedings 15}.\hskip 1em plus 0.5em
  minus 0.4em\relax Springer, 2003, pp. 1--13.

\bibitem{yang2004trangen}
K.~Yang, K.-T. Cheng, and L.-C. Wang, ``Trangen: A sat-based atpg for
  path-oriented transition faults,'' in \emph{ASP-DAC 2004: Asia and South
  Pacific Design Automation Conference 2004 (IEEE Cat. No. 04EX753)}.\hskip 1em
  plus 0.5em minus 0.4em\relax IEEE, 2004, pp. 92--97.

\bibitem{sorensson2005minisat}
N.~Sorensson and N.~Een, ``Minisat v1. 13-a sat solver with conflict-clause
  minimization,'' \emph{SAT}, vol. 2005, no.~53, pp. 1--2, 2005.

\bibitem{BiereFazekasFleuryHeisinger}
A.~Biere, K.~Fazekas, M.~Fleury, and M.~Heisinger, ``{CaDiCaL}, {Kissat},
  {Paracooba}, {Plingeling} and {Treengeling} entering the {SAT Competition
  2020},'' in \emph{Proc.~of {SAT Competition} 2020 -- Solver and Benchmark
  Descriptions}, ser. Department of Computer Science Report Series B, T.~Balyo,
  N.~Froleyks, M.~Heule, M.~Iser, M.~J{\"a}rvisalo, and M.~Suda, Eds., vol.
  B-2020-1.\hskip 1em plus 0.5em minus 0.4em\relax University of Helsinki,
  2020, pp. 51--53.

\bibitem{selman1993local}
B.~Selman, H.~A. Kautz, B.~Cohen \emph{et~al.}, ``Local search strategies for
  satisfiability testing.'' \emph{Cliques, coloring, and satisfiability},
  vol.~26, pp. 521--532, 1993.

\bibitem{cadical}
M.~Osama and A.~Wijs, ``Parafrost at the sat race 2021,'' \emph{SAT COMPETITION
  2021}, vol.~32, 2021.

\bibitem{queue2019cadical}
S.~D. QUEUE, ``Cadical at the sat race 2019,'' \emph{SAT RACE 2019}, p.~8,
  2019.

\bibitem{fleury2020cadical}
A.~Fleury and M.~Heisinger, ``Cadical, kissat, paracooba, plingeling and
  treengeling entering the sat competition 2020,'' \emph{SAT COMPETITION}, vol.
  2020, p.~50, 2020.

\bibitem{audemard2018glucose}
G.~Audemard and L.~Simon, ``On the glucose sat solver,'' \emph{International
  Journal on Artificial Intelligence Tools}, vol.~27, no.~01, p. 1840001, 2018.

\bibitem{lu2003circuit}
F.~Lu, L.-C. Wang, K.-T. Cheng, and R.-Y. Huang, ``A circuit sat solver with
  signal correlation guided learning,'' in \emph{2003 Design, Automation and
  Test in Europe Conference and Exhibition}.\hskip 1em plus 0.5em minus
  0.4em\relax IEEE, 2003, pp. 892--897.

\bibitem{audemard2009glucose}
G.~Audemard and L.~Simon, ``Glucose: a solver that predicts learnt clauses
  quality,'' \emph{SAT Competition}, pp. 7--8, 2009.

\bibitem{huang2022neural}
J.~Huang, H.-L. Zhen, N.~Wang, H.~Mao, M.~Yuan, and Y.~Huang, ``Neural fault
  analysis for sat-based atpg,'' in \emph{2022 IEEE International Test
  Conference (ITC)}.\hskip 1em plus 0.5em minus 0.4em\relax IEEE, 2022, pp.
  36--45.

\bibitem{gagliolo2010algorithm}
M.~Gagliolo and J.~Schmidhuber, ``Algorithm selection as a bandit problem with
  unbounded losses,'' in \emph{Learning and Intelligent Optimization: 4th
  International Conference, LION 4, Venice, Italy, January 18-22, 2010.
  Selected Papers 4}.\hskip 1em plus 0.5em minus 0.4em\relax Springer, 2010,
  pp. 82--96.

\bibitem{goldberg2003verification}
E.~Goldberg and Y.~Novikov, ``Verification of proofs of unsatisfiability for
  cnf formulas,'' in \emph{2003 Design, Automation and Test in Europe
  Conference and Exhibition}.\hskip 1em plus 0.5em minus 0.4em\relax IEEE,
  2003, pp. 886--891.

\bibitem{mishchenko2006improvements}
A.~Mishchenko, S.~Chatterjee, R.~Brayton, and N.~Een, ``Improvements to
  combinational equivalence checking,'' in \emph{Proceedings of the 2006
  IEEE/ACM international conference on Computer-aided design}, 2006, pp.
  836--843.

\bibitem{liang1995identifying}
H.-C. Liang, C.~L. Lee, and J.~E. Chen, ``Identifying untestable faults in
  sequential circuits,'' \emph{IEEE Design \& Test of computers}, vol.~12,
  no.~03, pp. 14--23, 1995.

\bibitem{heragu1997fast}
K.~Heragu, J.~H. Patel, and V.~D. Agrawal, ``Fast identification of untestable
  delay faults using implications,'' in \emph{iccad}, 1997, pp. 642--647.

\bibitem{selsam2019guiding}
D.~Selsam and N.~Bj{\o}rner, ``Guiding high-performance sat solvers with
  unsat-core predictions,'' in \emph{International Conference on Theory and
  Applications of Satisfiability Testing}.\hskip 1em plus 0.5em minus
  0.4em\relax Springer, 2019, pp. 336--353.

\bibitem{moskewicz2001chaff}
M.~W. Moskewicz, C.~F. Madigan, Y.~Zhao, L.~Zhang, and S.~Malik, ``Chaff:
  Engineering an efficient sat solver,'' in \emph{Proceedings of the 38th
  annual Design Automation Conference}, 2001, pp. 530--535.

\bibitem{guo2022machine}
W.~Guo, J.~Yan, H.-L. Zhen, X.~Li, M.~Yuan, and Y.~Jin, ``Machine learning
  methods in solving the boolean satisfiability problem,'' \emph{arXiv preprint
  arXiv:2203.04755}, 2022.

\bibitem{selsam2018learning}
D.~Selsam, M.~Lamm, B.~B{\"u}nz, P.~Liang, L.~de~Moura, and D.~L. Dill,
  ``Learning a sat solver from single-bit supervision,'' \emph{arXiv preprint
  arXiv:1802.03685}, 2018.

\bibitem{amizadeh2018learning}
S.~Amizadeh, S.~Matusevych, and M.~Weimer, ``Learning to solve circuit-sat: An
  unsupervised differentiable approach,'' in \emph{International Conference on
  Learning Representations}, 2018.

\bibitem{li2022deepsat}
M.~Li, Z.~Shi, Q.~Lai, S.~Khan, and Q.~Xu, ``Deepsat: An eda-driven learning
  framework for sat,'' \emph{arXiv preprint arXiv:2205.13745}, 2022.

\bibitem{zhang2020nlocalsat}
W.~Zhang, Z.~Sun, Q.~Zhu, G.~Li, S.~Cai, Y.~Xiong, and L.~Zhang, ``Nlocalsat:
  Boosting local search with solution prediction,'' \emph{arXiv preprint
  arXiv:2001.09398}, 2020.

\bibitem{kurin2020can}
V.~Kurin, S.~Godil, S.~Whiteson, and B.~Catanzaro, ``Can q-learning with graph
  networks learn a generalizable branching heuristic for a sat solver?''
  \emph{Advances in Neural Information Processing Systems}, vol.~33, pp.
  9608--9621, 2020.

\bibitem{wang2021neurocomb}
W.~Wang, Y.~Hu, M.~Tiwari, S.~Khurshid, K.~McMillan, and R.~Miikkulainen,
  ``Neurocomb: Improving sat solving with graph neural networks,'' \emph{arXiv
  e-prints}, pp. arXiv--2110, 2021.

\bibitem{vaswani2017attention}
A.~Vaswani, N.~Shazeer, N.~Parmar, J.~Uszkoreit, L.~Jones, A.~N. Gomez,
  {\L}.~Kaiser, and I.~Polosukhin, ``Attention is all you need,''
  \emph{Advances in neural information processing systems}, vol.~30, 2017.

\bibitem{meng2020readnet}
C.~Meng, M.~Chen, J.~Mao, and J.~Neville, ``Readnet: A hierarchical transformer
  framework for web article readability analysis,'' in \emph{European
  Conference on Information Retrieval}.\hskip 1em plus 0.5em minus 0.4em\relax
  Springer, 2020, pp. 33--49.

\bibitem{wang2019learning}
Q.~Wang, B.~Li, T.~Xiao, J.~Zhu, C.~Li, D.~F. Wong, and L.~S. Chao, ``Learning
  deep transformer models for machine translation,'' \emph{arXiv preprint
  arXiv:1906.01787}, 2019.

\bibitem{dosovitskiy2020image}
A.~Dosovitskiy, L.~Beyer, A.~Kolesnikov, D.~Weissenborn, X.~Zhai,
  T.~Unterthiner, M.~Dehghani, M.~Minderer, G.~Heigold, S.~Gelly \emph{et~al.},
  ``An image is worth 16x16 words: Transformers for image recognition at
  scale,'' \emph{arXiv preprint arXiv:2010.11929}, 2020.

\bibitem{liu2021swin}
Z.~Liu, Y.~Lin, Y.~Cao, H.~Hu, Y.~Wei, Z.~Zhang, S.~Lin, and B.~Guo, ``Swin
  transformer: Hierarchical vision transformer using shifted windows,'' in
  \emph{Proceedings of the IEEE/CVF International Conference on Computer
  Vision}, 2021, pp. 10\,012--10\,022.

\bibitem{shi2021transformer}
F.~Shi, C.~Lee, M.~K. Bashar, N.~Shukla, S.-C. Zhu, and V.~Narayanan,
  ``Transformer-based machine learning for fast sat solvers and logic
  synthesis,'' \emph{arXiv preprint arXiv:2107.07116}, 2021.

\bibitem{tseitin1983complexity}
G.~S. Tseitin, ``On the complexity of derivation in propositional calculus,''
  in \emph{Automation of reasoning}.\hskip 1em plus 0.5em minus 0.4em\relax
  Springer, 1983, pp. 466--483.

\bibitem{caruana1997multitask}
R.~Caruana, ``Multitask learning,'' \emph{Machine learning}, vol.~28, no.~1,
  pp. 41--75, 1997.

\bibitem{kullback1951information}
S.~Kullback and R.~A. Leibler, ``On information and sufficiency,'' \emph{The
  annals of mathematical statistics}, vol.~22, no.~1, pp. 79--86, 1951.

\bibitem{lin2017feature}
T.-Y. Lin, P.~Doll{\'a}r, R.~Girshick, K.~He, B.~Hariharan, and S.~Belongie,
  ``Feature pyramid networks for object detection,'' in \emph{Proceedings of
  the IEEE conference on computer vision and pattern recognition}, 2017, pp.
  2117--2125.

\bibitem{marques2021conflict}
J.~Marques-Silva, I.~Lynce, and S.~Malik, ``Conflict-driven clause learning sat
  solvers,'' in \emph{Handbook of satisfiability}.\hskip 1em plus 0.5em minus
  0.4em\relax ios Press, 2021, pp. 133--182.

\bibitem{kingma2014adam}
D.~P. Kingma and J.~Ba, ``Adam: A method for stochastic optimization,''
  \emph{arXiv preprint arXiv:1412.6980}, 2014.

\bibitem{ozolins2021goal}
E.~Ozolins, K.~Freivalds, A.~Draguns, E.~Gaile, R.~Zakovskis, and S.~Kozlovics,
  ``Goal-aware neural sat solver,'' \emph{arXiv preprint arXiv:2106.07162},
  2021.

\bibitem{oh2015between}
C.~Oh, ``Between sat and unsat: the fundamental difference in cdcl sat,'' in
  \emph{Theory and Applications of Satisfiability Testing--SAT 2015: 18th
  International Conference, Austin, TX, USA, September 24-27, 2015, Proceedings
  18}.\hskip 1em plus 0.5em minus 0.4em\relax Springer, 2015, pp. 307--323.

\bibitem{nadel2010boosting}
A.~Nadel, ``Boosting minimal unsatisfiable core extraction,'' in \emph{Formal
  Methods in Computer Aided Design}.\hskip 1em plus 0.5em minus 0.4em\relax
  IEEE, 2010, pp. 221--229.

\end{thebibliography}

\end{document}